\documentclass{ifcolog}

\usepackage{url}
\newtheorem{definition}{Definition}
\usepackage{subcaption}
\usepackage[most]{tcolorbox}

\colorlet{punct}{red!60!black}
\definecolor{background}{HTML}{FFFFFF}
\definecolor{delim}{RGB}{20,105,176}
\colorlet{numb}{magenta!60!black}

\usepackage{listings}
\lstdefinelanguage{json}{
basicstyle=\small\ttfamily,
numbers=none,
numberstyle=\scriptsize,
stepnumber=1,
numbersep=8pt,
showstringspaces=false,
breaklines=true,
backgroundcolor=\color{background},
literate=
*{0}{{{\color{numb}0}}}{1}
{1}{{{\color{numb}1}}}{1}
{2}{{{\color{numb}2}}}{1}
{3}{{{\color{numb}3}}}{1}
{4}{{{\color{numb}4}}}{1}
{5}{{{\color{numb}5}}}{1}
{6}{{{\color{numb}6}}}{1}
{7}{{{\color{numb}7}}}{1}
{8}{{{\color{numb}8}}}{1}
{9}{{{\color{numb}9}}}{1}
{:}{{{\color{punct}{:}}}}{1}
{,}{{{\color{punct}{,}}}}{1}
{\{}{{{\color{delim}{\{}}}}{1}
{\}}{{{\color{delim}{\}}}}}{1}
{[}{{{\color{delim}{[}}}}{1}
{]}{{{\color{delim}{]}}}}{1},
}

\title{A Formalization of Kant's Second Formulation of the Categorical Imperative}
\titlerunning{Kant's Second Formulation of the Categorical Imperative}
\titlethanks{We would like to thank the three anonymous reviewers for their valuable comments that helped us to improve the paper.}
\addauthor[lindner@informatik.uni-freiburg.de]{Felix Lindner}{Foundations of Artificial Intelligence, University of Freiburg, Germany}
\addauthor[mmbe@dtu.dk]
{Martin Mose Bentzen}
{Management Engineering, Technical University of Denmark, Lyngby, Denmark}
\authorrunning{Lindner and Bentzen}
\begin{document}
\maketitle
\begin{abstract}
We present a formalization and computational implementation of the second formulation of Kant's categorical imperative. This ethical principle requires an agent to never treat someone merely as a means but always also as an end. Here we interpret this principle in terms of how persons are causally affected by actions. We introduce Kantian causal agency models in which moral patients, actions, goals, and causal influence are represented, and we show how to formalize several readings of Kant's categorical imperative that correspond to Kant's concept of strict and wide duties towards oneself and others. Stricter versions handle cases where an action directly causally affects oneself or others, whereas the wide version maximizes the number of persons being treated as an end. We discuss limitations of our formalization by pointing to one of Kant's cases that the machinery cannot handle in a satisfying way.
\end{abstract}

\section{Introduction}

It has been suggested that artificial agents, such as social robots and software bots, must be programmed in an ethical way in order to remain beneficial to human beings. One prominent ethical theory was proposed by Immanuel Kant \cite{Kan85}. Here, we propose a formalization and implementation of Kant's ethics with the purpose of guiding artificial agents that are to function ethically. In particular, the system will be able to judge whether actions are ethically permissible according to Kant's ethics. In order to accomplish this we focus on the second formulation of Kant's categorical imperative. Kant proposed three formulations of the categorical imperative. We formalize and implement the second formulation and do not take a stance on the interrelation of Kant's three formulations.
The second formulation of Kant's categorical imperative reads:

\begin{quote}
Act in such a way that you treat humanity, whether in your own person or in the person of any other, never merely as a means to an end, but always at the same time as an end. (Kant, 1785)
\end{quote}

We take it to be the core of the second formulation of the categorical imperative that all rational beings affected by our actions must be considered as part of the goal of the action.

The paper is structured as follows: We first briefly review related work. Then, building upon our earlier work \cite{LBN17}, we introduce an extension of Pearl-Halpern-style causal networks which we call Kantian causal agency models. These models serve as a formal apparatus to model the morally relevant aspects of situations. We then define an action's permissibility due to the categorical imperative, while considering two readings of \emph{being treated as a means}. To deal with Kant's wider duties, we introduce an extra condition according to which an agent should maximize the number of persons being treated as an end. Finally, we briefly showcase the computational implementation of the categorical imperative within the HERA software library\footnote{\url{http://www.hera-project.com}}.

\section{Related work}
In machine ethics, several ethical theories have been formalized and
implemented, e.g., utilitarianism, see \cite{Hor01,AP05}, the principle of
double effect, see \cite{Ben16,GB17}, pareto permissibility, see
\cite{LBN17}, and Asimov's laws of robotics, see \cite{Win14}.

It has been suggested for some time that Kant's ethics could be
formalized and implemented computationally, see \cite{Pow06,Abn12}. Powers \cite{Pow06} suggests
three possible ways of formalizing Kant's first formulation of the
categorical imperative, through deontic logic, non-monotonic logic, or
belief revision. The first formulation of the categorical imperative
states that you must be able to want that the subjective reasoning (or
maxim) motivating your action becomes a universal law and as Kant claims
that this in some cases is a purely formal matter, it should be possible
to formalize it. However, Powers does not provide details of a
formalization or a computational implementation, so the formalization of
the first formulation in effect remains an open problem.

The work presented here differs in that we focus on the second
formulation of the categorical imperative and in that we present a
precise formal representation and computational implementation of the
formal theory. Rather than taking a starting point in one of the
paradigms Powers suggests, we use formal semantics and causal agency
modelling as this is fitting for the means-end reasoning central to the
second formulation. Philosophically, our formalization is best seen as a rational reconstruction within this framework of what we take to be the central ideas of Kant's second formulation.

We think the second formulation has some intuitive appeal also to modern people, many people perceive that there is something morally wrong in using people (including yourself) without consideration of how it affects them. Ultimately, although we are sensitive to Kant's original text, the goal of our work is not to get close to a correct interpretation of Kant, but to show that our interpretation of Kant's ideas can contribute to the development of machine ethics. To meet this goal, our interpretation has to be detailed and explicit enough to provide a decision mechanism for the permissibility or not of specific actions in specific situations.

\section{Kantian causal agency models}
\label{sect:causal_agency_models}

In order to formalize the second formulation of the categorical imperative, we assume some background theory. First, we assume that \emph{actions} are performed by \emph{agents}, and that actions and their \emph{consequences} can \emph{affect} a set of \emph{moral patients}, i.e. persons who must be considered ethically in a situation. The agent itself is also one of the moral patients. The agent has available a set of actions which will have consequences given \emph{background conditions}. Some of the action's consequences are the \emph{goals} of the action. The actions and consequences that together cause my goal are the \emph{means} of the action. Patients, who are affected by these means are \emph{treated as a means}, and patients, who are affected by my goal are \emph{treated as an end}. For example, I (agent) have the option available to press the light switch (action), and given that the light bulb is not broken (background condition), the light will go on (consequence), which leads to me being able to read my book (consequence). The last consequence was also my goal, and it affects me in a positive way. The action thus treats me as an end.

Within this informally characterized framework, we can reformulate the second formulation of the categorical imperative as follows:

\begin{quote}
Act in such a way, that whoever is treated as a means through your action (positively or negatively and including yourself), must also be treated as an end of your action.
\end{quote}

The purpose of what follows is to formalize these intuitions. As a first step, we now give the formal definition of the models we will be using in Definition \ref{def:kantian_causal_agency_model}. We call these models \emph{Kantian causal agency models} to set them apart from the causal agency models we used in our earlier work \cite{LBN17}, and which had no formal tools to consider moral patients affected by one's actions.

\begin{definition}[Kantian Causal Agency Model]
\label{def:kantian_causal_agency_model}
A \emph{Kantian causal agency model} $M$ is a tuple $(A, B, C, F, G, P, K, W)$, where $A$ is the set of \emph{action variables}, $B$ is a set of \emph{background variables}, $C$ is a set of \emph{consequence variables}, $F$ is a set of modifiable \emph{boolean structural equations}, $G = (Goal_{a_1}, \ldots, Goal_{a_n})$ is a list of sets of variables (one for each action), $P$ is a set of moral patients (includes a name for the agent itself), $K$ is the ternary \emph{affect relation} $K \subseteq (A\cup C)\times P \times \{+,-\}$, and $W$ is a set of \emph{interpretations} (i.e., truth assignments) over $A\cup B$.

\end{definition}

$A$ (actions), $B$ (background variables) and $C$ (consequences) are finite sets of boolean variables with B and C possibly empty. $W$ is a set of boolean interpretations of $A\cup B$. Thus, the elements of $W$ set the truth values of those variables that are determined externally, and thus specify the concrete situation. We require that all interpretations in $W$ assign true to exactly one action $a\in A$. As a notational convention, by $M, w_a$ and $M, w_b$ we distinguish two situations that only differ in that in the first situation, action $a$ is performed, and in the second situation, action $b$ is performed.

Causal influence is determined by the set $F$ of boolean-valued structural equations. Each variable $c_i\in C$ is associated with the function $f_i\in F$. This function will give $c_i$ its value under an interpretation $w\in W$. An interpretation $w$ is extended to the consequence variables as follows: For a variable $c_i\in C$, let $\{c_{i1},\ldots, c_{im-1}\}$ be the variables of $C\setminus\{c_i\}$, $B = {b_1, \ldots, b_k}$, and $A = \{a_1, \ldots, a_n\}$ the action variables.
The assignment of truth values to consequences is determined by:
\begin{eqnarray*}
w(c_i) &=& f_i(w(a_1),\ldots, w(a_n),
w(b_1), \ldots, w(b_k),
w(c_{i1}),\ldots, w(c_{im-1}))
\end{eqnarray*}

To improve readability, we will use the notation $c := \phi$ to express that $c$ is true if $\phi$ is true, where $\phi$ can be any boolean formula containing variables from $A\cup B\cup C$ and its negations. For instance, the boolean structural equations for the light-switch example will be written as
$F = \{lightOn := press \land \lnot bulbBroken, canReadBook := lightOn\}$.

In the general setting, it may be unfeasible to extend an interpretation from the action variables to the rest of the variables, because it is possible that the value of some variable depends on the value of another variable, and the value of the latter variable depends on the value of the former. Dependence is defined in Definition \ref{def:dependence}.

\begin{definition}[Dependence]
\label{def:dependence}
Let $v_i\in C, v_j \in A\cup B\cup C$ be distinct variables. The variable $v_i$ \emph{depends on} variable $v_j$, if,
for some vector of boolean values, $f_i(\ldots, v_j = 0, \ldots) \neq f_i(\ldots, v_j = 1, \ldots)$.
\end{definition}

Following Halpern \cite{Hal16}, we restrict causal agency models to acyclic models, i.e., models in which no two variables are mutually dependent on each other. First, note that the values of action variables in set $A$ and the values of background variables in set $B$ are determined externally by the interpretations in $W$. Thus, the truth values of action variables and background variables do not depend on any other variables. Additionally, we require that the transitive closure, $\prec$, of the dependence relation is a partial order on the set of variables: $v_1\prec v_2$ reads ``$v_1$ is causally modified by $v_2$''. This enforces absence of cycles. In case of acyclic models, the values of all consequence variables can be determined unambiguously: First, there will be consequence variables only causally modified by action and/or background variables, and whose truth value can thus be determined by the values set by the interpretation. Call these consequence variables \emph{level one}. On \emph{level two}, there will be consequence variables causally modified by action variable, background variables, and level-one consequence variables, and so on \cite{Ben16,Hal16}.

Some of the definitions below will make use of causality. Thus, to take causation into account, Definition \ref{def:but-for} defines the relation of $y$ being a but-for cause of $\phi$, see \cite{Hal16}.
Definition \ref{def:but-for} makes use of \emph{external interventions} on models. An external interventions $X$ consists of a set of literals (viz., action variables, consequence variables, background variables, and negations thereof). Applying an external intervention to a causal agency model results in a new causal agency model $M_X$. The truth of a variable $v\in A\cup B\cup C$ in $M_X$ is determined in the following way: If $v \in X$, then $v$ is true in $M_X$, if $\lnot v\in X$, then $v$ is false in $M_X$, and if neither $v\in X$ nor $\lnot v\in X$, then the truth of $v$ is determined according to its structural equation in $M$. External interventions thus override structural equations of the variables occurring in $X$.

\begin{definition}[Actual but-for cause]
\label{def:but-for}
Let $y$ be a literal and $\phi$ a formula. We say that $y$ is an \emph{actual but-for cause} of $\phi$ (notation: $y \leadsto \phi$) in the situation the agent choses option $w_a$ in model $M_X$, if and only if $M_X, w_a\models y\land \phi$ and $M_{(X\setminus\{y\})\cup\{\lnot y\}},w_a\models \lnot \phi$.
\end{definition}

The first condition requires that both the cause and the effect must be actual. The second condition requires that if $y$ had not been the case, then $\phi$ would have not occurred. Thus, in the chosen situation, $y$ was necessary to bring about $\phi$. Consider again the book reading situation $M, w$, such that $w(bulbBroken) = \bot, w(press) = \top$. Due to the structural equations (see above), we have both $M, w \models press$ and $M, w \models canReadBook$. Also, in the intervention where the agent does not press the light switch, the agent cannot read the book, $M_{\{\lnot press\}} \models \lnot canReadBook$. Therefore, in situation $M, w$, $press$ is a but-for cause of $canReadBook$.

Generally, the definition of but-for cause allows to talk about individual actions and consequences and their causal effects on other individual consequences in the given situation, as well as counterfactual effects if the situation were different from the actual situation. The definition does not allow conjunctive or disjunctive causes.
Consequently, this definition of causality does not cope with cases of preemption. For instance, consider the agent shoots at someone who is already about to die, because he was poisened just a minute ago. In this case, the agent's shot is not a but-for cause for the patient's death---but the disjunction of the agent's shot and the patient being poisened is. Our examples work with the simpler but-for causality, so we do not discuss more sophisticated definitions of causality (but see \cite{Hal16}).

Based on the concept of but-for cause the useful concept of \emph{direct consequences} is introduced via Definition \ref{def:direct-consequence}.
\begin{definition}[Direct Consequence]
\label{def:direct-consequence}
A consequence $c \in C$ is a \emph{direct consequence} of $v \in A\cup B\cup C$ in the situation $M_X, w_a$ iff $M_X, w_a\models v \leadsto c$.
\end{definition}

With regard to modeling moral patients affected by effects, we assert that persons can be affected by actions or consequences either in a positive or in a negative way. To represent that some action or consequence (knowingly) affects a person positively or negatively, we introduce the notations $\triangleright_{+}$ and $\triangleright_{-}$, respectively. Thus, $M_X, w_a \models c \triangleright_{+} p$ holds iff $(a, c, +) \in K$, and $M_X, w_a \models c \triangleright_{-} p$ holds iff $(a, c, -) \in K$. We use $\triangleright$ in case the valence of affection is not relevant.
As a means to refer to the goals of some action, we define $M_X, w_a \models Goal(c)$ iff $c \in Goal_a$, i.e., a consequence $c$ is the goal in the agent's chosen situation $w_a$ iff $c$ is in the set of goals associated with action $a$ (cf., Def.\ \ref{def:kantian_causal_agency_model}).

This finalizes the exposition of the background theory.

\section{Categorical imperative defined}
\label{sect:strict_definition}

We now consider how to make permissibility judgments about actions as defined in the context of Kantian causal agency models using the categorical imperative.
The second formulation of the categorical imperative requires an agent to never treat someone merely as a means but always also as an end.
Thus, to formalize under which conditions an action is permitted by the categorical imperative, we first define the concept of someone being \emph{treated as an end} (Definition \ref{def:treated_as_end}). We then proceed to formalize two possible readings of the concept of someone being \emph{treated as a means} (Definition \ref{def:treated_as_means1} and Definition \ref{def:treated_as_means2}).

\begin{definition}[Treated as an End]
\label{def:treated_as_end}
A patient $p\in P$ is \emph{treated as an end} by action $a$, written $M_X, w_a \models End(p)$, iff the following conditions hold:
\begin{enumerate}
\item Some goal $g$ of $a$ affects $p$ positively. \\ $M_X, w_a \models \bigvee_g\big(Goal(g) \land g \triangleright_{+} p\big)$.
\item None of the goals of $a$ affect $p$ negatively.\\
$M_X, w_a \models \bigwedge_g (Goal(g) \rightarrow \lnot (g \triangleright_{-} p))$
\end{enumerate}
\end{definition}

Thus, being treated as an end by some action means that some goal of the action affects one in a positive way. One could say that the agent of the action, by performing that action, considers those who benefit from her goal.
Things are less clear regarding the concept `being treated as a means'. As a first step, we define two versions of the concept which we refer to as \emph{Reading 1} and \emph{Reading 2}. Both readings make use of the causal consequences of actions. Reading 1 considers a person used as a means in case she is affected by some event that causally brings about some goal of the action.
\begin{definition}[Treated as a Means, Reading 1]
\label{def:treated_as_means1}
A patient $p\in P$ is \emph{treated as a means} by action $a$ (according to Reading 1), written $M_X, w_a \models Means_1(p)$, iff there is some $v \in A\cup C$, such that $v$ affects $p$, and $v$ is a cause of some goal $g$, i.e., $M_X, w_a \models \bigvee_v \big((a\leadsto v \land v\triangleright p) \land \bigvee_g (v \leadsto g \land Goal(g))\big)$.
\end{definition}

As a consequence, negative side effects are permitted under Reading 1. Consider, for instance, the classical trolley dilemma, where the agent has the choice to either pull the lever to lead the tram onto the second track killing one person, or refraining from pulling letting the tram kill five persons on the first track (see Fig.\ \ref{fig:model:trolley}). Under Reading 1, in case of pulling, the one agent---person 6 in Fig.\ \ref{fig:model:trolley}---is not treated as a means: If, counterfactually, person 6 survived although the switch was pulled, then this would not deactivate any of the agent's goals. Therefore, the formula $\lnot survive6 \leadsto survive1 \lor \ldots \lor \lnot survive6 \leadsto survive5$ is not satisfied by the model of the classical trolley problem. Reading 1 is probably closest to what we informally mean by `being treated as a means'.
\begin{figure*}[t!]
\centering
\begin{eqnarray*}
A &=& \{pull\}\\
C &=& \{survive1, \ldots, survive6\}\\
P &=& \{person1, \ldots, person6\}\\
F &=& \{survive1 := pull, \ldots, survive5 := pull, survive6 := \lnot pull\}\\
K &=& \{(survive1,person1,+), (\lnot survive1, person1, -), \ldots\}\\
G &=& (Goal_{pull} = \{survive1, \ldots, survive5\})
\end{eqnarray*}
\caption{Model of the classical trolley problem. Person 1 to person 5 are together on the one track, person 6 alone on the other track.}
\label{fig:model:trolley}
\end{figure*}

Reading 2 requires that everybody affected by any direct consequence of the action is considered as a goal.
\begin{definition}[Treated as a Means, Reading 2]
\label{def:treated_as_means2}
A patient $p\in P$ is \emph{treated as a means} by action $a$ (according to Reading 2), written $M_X, w_a \models Means_2(p)$, iff there is some direct consequence $v \in A\cup C$ of $a$, such that $v$ affects $p$, i.e., $M_X, w_a \models \bigvee_v \big(a\leadsto v \land v\triangleright p\big)$.
\end{definition}

Hence, under Reading 2, also the person on the second track must be considered as a goal. Consequently, everyone treated as a means according to Reading 1 is also treated as a means according Reading 2, but Reading 2 may include additional patients. Reading 2 is further from the everyday understanding of means-end reasoning, but is probably closer to what some people expect of a Kantian ethics, viz., that everyone affected by the direct consequences of one's actions must be considered. We consider it a feature of a formal framework that this distinction can be formalized, but we leave it for the modeler to decide which one of the readings is more useful for a given application.
One thing to note is that Kantian causal-agency models are meant to represent what an agent considers possible. Hence, the agent uses some patient as a means in case she knowingly affects that patient. Thus, the formalization does not require an agent to consider affected moral patients she was not aware of. For instance, if the reader of this paper feels affected by what she reads, then, of course, the authors are not using her as a means.

Having defined both \emph{being treated as an end} and \emph{being treated as a means}, the permissibility of actions according to the second formulation of the categorical imperative can now be defined in Definition \ref{def:categorical_imperative}. The formulation requires that no-one is merely used as a means, but always at the same time as an end.
\begin{definition}[Categorical Imperative]
\label{def:categorical_imperative}
An action $a$ is permitted according to the categorical imperative, iff for any $p\in P$, if $p$ is treated as a means (according to Reading $N$) then $p$ is treated as an end $M_X, w_a \models \bigwedge_{p\in P} (Means_N(p) \rightarrow End(p))$.
\end{definition}

There are thus two main reasons why an action is not permitted: Either a patient is treated as a means but is left out of consideration by the end of the action. Or, the action is done for an end that affects someone negatively.

\section{Cases of strict duty}

We will now provide examples that highlight aspects of the definition of the categorical imperative. Although these do not prove it correct in any formal sense they can be used to discuss its appeal as an ethical principle as an explication of Kant's ideas.
First, we rephrase three cases that contain what Kant calls strict duties (and two of which Kant himself used to explain his ideas).

\subsection{Example 1: Suicide}
\label{subsubsection:suicide}

Bob wants to commit suicide, because he feels so much pain he wants to be relieved from. This case can be modeled by a causal agency model $M_1$ that contains one action variable $suicide$ and a consequence variable $dead$, see Figure \ref{fig:models:m1m1}. Death is the goal of the suicide action (as modeled by the set $G$), and the suicide affects Bob (as modeled by the set $K$). In this case, it does not make a difference whether the suicide action affects Bob positively or negatively.

\begin{figure*}[t!]
\centering
\begin{subfigure}[t]{0.49\textwidth}
\begin{eqnarray*}
A &=& \{suicide\}\\
C &=& \{dead\}\\
P &=& \{Bob\}\\
F &=& \{dead := suicide\}\\
K &=& \{(suicide, Bob,+)\}\\
G &=& (Goal_{suicide} = \{dead\})\\
& &\phantom{\mspace{10mu}(survives, Bob,+)\}}
\end{eqnarray*}
\caption{Model $M_1$}
\label{fig:models:m1m1}
\end{subfigure}
\begin{subfigure}[t]{0.49\textwidth}
\begin{eqnarray*}
A &=& \{amputate\}\\
C &=& \{survives\}\\
P &=& \{Bob\}\\
F &=& \{survives := amputate\}\\
K &=& \{(amputate, Bob, -), \\& &\mspace{10mu}(survives, Bob,+)\}\\
G &=& (Goal_{amputate} = \{survives\})
\end{eqnarray*}
\caption{Model $M_1^*$}
\label{fig:models:m1m1s}
\end{subfigure}
\caption{Kantian Causal Agency Models yielding the impermissibility of Suicide ($M_1$) and the permissibility of Amputation ($M_1^*$).}
\label{fig:models:m1}
\end{figure*}

The model assumes that the suicide affects no-one other than Bob, because Kant's argument is not about the effect of suicide on other people but about the lack of respect of the person committing suicide. The reason why Bob's suicide is not permitted is that the person affected by the suicide, viz., Bob, does not benefit from the goal, because he is destroyed and thus cannot be affected positively by it. He is thus treated as a means to his own annihilation from which he receives no advancement. Therefore, the first condition of the categorical imperative (Definition \ref{def:categorical_imperative}) is violated according to both readings (1 and 2), because $M_1, w_{suicide} \models Means_{\{1,2\}}(Bob)$ holds but $M_1, w_{suicide} \models End(Bob)$ does not.

As noted above, it could also be said that the suicide affects Bob negatively, and the action would also be impermissible. The reason for the impermissibility of suicide also in this case is not due to the fact that Bob does something harmful towards himself. As Kant also remarks, other harmful actions would be allowed, e.g., risking your life or amputating a leg to survive. To see this, consider Fig.\ \ref{fig:models:m1m1s}, where $M_1$ has been be slightly modified to $M_1^*$: Rename $suicide$ to $amputate$ and $dead$ to $survives$. Moreover, add $(amputate, Bob, -)$ to $K$. In this case, Bob is positively affected by the goal, and thus the act of amputation is permitted. The modified example also shows that in some cases, the categorical imperative is more permissive than the principle of double effect, which strictly speaking never allows negative means to an end (cf., \cite{Ben16}).

\subsection{Example 2: Giving flowers}

The fact that an action can be judged as impermissible by the categorical imperative although no-one is negatively affected is a property of the categorical imperative that inheres in no other moral principles formalized so far. The following example showcases another situation to highlight this property: Bob gives Alice flowers in order to make Celia happy when she sees that Alice is thrilled about the flowers. Alice being happy is not part of the goal of Bob's action.
We model this case by considering the Kantian causal agency model $M_2$ shown in Figure \ref{fig:models:m2m2}.
\begin{figure*}[t!]
\centering
\begin{subfigure}[t]{0.49\textwidth}
\centering
\begin{eqnarray*}
A &=& \{give\_flowers\}\\
C &=& \{alice\_happy, celia\_happy\}\\
P &=& \{Bob, Alice, Celia\}\\
F &=& \{ alice\_happy := give\_flowers\\
& & \mspace{9mu} celia\_happy := alice\_happy\}\\
K &=& \{(alice\_happy, Alice, +),\\&&(celia\_happy, Celia, +)\}\\
G &=& (Goal_{give\_flowers} = \\&&\mspace{50mu}\{celia\_happy\})
\end{eqnarray*}
\vspace{0.3cm}
\caption{Model $M_2$}
\label{fig:models:m2m2}
\end{subfigure}
\begin{subfigure}[t]{0.49\textwidth}
\begin{eqnarray*}
A &=& \{give\_flowers\}\\
C &=& \{alice\_happy, celia\_happy\}\\
P &=& \{Bob, Alice, Celia\}\\
F &=& \{ alice\_happy := give\_flowers\\
& & \mspace{9mu} celia\_happy := alice\_happy\}\\
K &=& \{(alice\_happy, Alice, +),\\&&(celia\_happy, Celia, +)\}\\
G &=& (Goal_{give\_flowers} = \\&&\mspace{5mu}\{celia\_happy, alice\_happy\})
\end{eqnarray*}
\caption{Model $M_2^*$}
\label{fig:models:m2m2s}
\end{subfigure}
\caption{Kantian Causal Agency Models yielding the impermissibility of giving flowers to Alice to make Celia happy ($M_2$) and the permissibility of doing so if making Celia happy is a goal as well ($M_2^*$).}
\label{fig:models:m2}
\end{figure*}

In the model $M_2$, the action $give\_flowers$ is not permitted according to the categorical imperative, because Bob is using Alice as a means to make Celia happy, but not considering her as part of the goal of the action. This action is immoral, even though the action has positive consequences for all, and no bad consequence are used to obtain a good one. Again, this example shows how the Kantian principle differs from other ethical principles such as utilitarianism and the principle of double effect, because these principles would permit the action.

The model $M_2$ can be extended to model $M_2^*$ shown in Figure \ref{fig:models:m2m2s}. In model $M_2^*$, Bob's action is permitted by the Kantian principle. The only thing in which $M_2^*$ differs from $M_2$ is that the variable $alice\_happy$ is added to the set $Goal_{give\_flowers}$. In this case, Alice is both treated as a means and treated as an end, which is permitted by the categorical imperative.

The flower example demonstrates how demanding the categorical imperative is, because the principles requires that everybody affected by ones' action must be treated as a goal: This includes the taxi driver that drives you to your destination, as well as the potential murderer you defend yourself against. In these examples, the ethical principle requires one to, e.g., have the taxi driver's earning money among one's goals, and the murderer's not going to jail.

\subsection{Example 3: False promise}

We return to a case mentioned by Kant himself. Consider that Bob makes a false promise to Alice. Bob borrows one 100 Dollars from Alice with the goal of keeping the money forever. He knows that it is an inevitable consequence of borrowing the money that he will never pay it back. Figure \ref{fig:model:m3m3} shows the model of this situation, $M_3$.
\begin{figure*}[t!]
\centering
\begin{eqnarray*}
A &=& \{borrow\}\\
C &=& \{bob\_keeps\_100Dollar\_forever\}\\
P &=& \{Alice\}\\
F &=& \{bob\_keeps\_100Dollar\_forever := borrow\}\\
K &=& \{(borrow,Bob,+), (borrow, Alice, -), \\
& & (bob\_keeps\_100Dollar\_forever, Bob,+), \\
& & (bob\_keeps\_100Dollar\_forever, Alice, -)\}\\
G &=& (Goal_{borrow} = \{bob\_keeps\_100Dollar\_forever\})
\end{eqnarray*}
\caption{Model $M_3$ for the case of Bob making a false promise to Alice.}
\label{fig:model:m3m3}
\end{figure*}
The action is impermissible, because Alice is treated as a means (by both Reading 1, Definition \ref{def:treated_as_means1}, and Reading 2, Definition \ref{def:treated_as_means2}). However, none of the two conditions for `being treated as an end' (Definition \ref{def:treated_as_end}) are met: None of the goals affects Alice positively, and Bob's goal affects her negatively.

\section{Cases of wide duty}
\label{sect:wide_definition}

Examples 1, 2 and 3 are instances of what Kant calls necessary, strict, narrower duties to oneself and to others, and it seems obvious they involve using a person as a means. Kant also presents two other examples to which we now turn in this section. These involve what Kant calls contingent, meritorious, or wider duties. His arguments for these appear more vague and at least from our perspective harder to handle.
We now turn to wide duties and discuss, through an example, how actions that indirectly affect others by refraining from preventing harmful consequences could be handled in the formal framework. Another example will demonstrate where the limitations of our formalization attempt are.

\subsection{Example 4: Not helping others}

Bob who has everything he needs, does not want to help Alice who is in need. Let us assume she is drowning and Bob is refraining from saving her live.
Formally, the situation in the example can be represented with a causal agency model $M_4$ that contains one background variable $accident$ representing the circumstances that led to Alice being in dire straits, two action variables $rescue$ and $\mathit{refrain}$ and a consequence variable $drown$. Moreover, $\lnot drown$ is the goal of $rescue$. See Figure \ref{fig:model:m4m4} for the specification of the model.
\begin{figure*}[t!]
\centering
\begin{eqnarray*}
A &=& \{rescue, \mathit{refrain}\}\\
B &=& \{accident\}\\
C &=& \{drown\}\\
P &=& \{Alice, Bob\}\\
F &=& \{drown := accident \land \lnot rescue\}\\
K &=& \{(drown,Alice,-), (\lnot drown, Alice, +)\}\\
G &=& (Goal_{rescue} = \{\lnot drown\}, Goal_{\mathit{refrain}} = \emptyset)
\end{eqnarray*}
\caption{Model $M_4$ for the impermissibility of not helping others.}
\label{fig:model:m4m4}
\end{figure*}

According to the categorical imperative using Readings 1 and 2 of `being treated as a means' both $rescue$ and $\mathit{refrain}$ are permitted. Bob is strictly speaking not using Alice as a means by going about his business. Kant gives us a clue of how to formalize an argument against refraining in that he says we have to make other people's ends our own as far as possible. Kant writes that `For a positive harmony with humanity as an end in itself, what is required is that everyone positively tries to further the ends of others as far as he can.' One way of understanding this is as an additional requirement on top of the categorical imperative of choosing an action whose goals affect most people positively. This understanding is captured in Definition \ref{def:meritorious}.
\begin{definition}[Meritorious principle]
\label{def:meritorious}
Among actions permitted by the categorical imperative, choose one whose goals affect most patients positively.
\end{definition}

The meritorious principle thus goes beyond simply avoiding to treat others as means by actively helping them. As formulated here, the principle is compatible with the categorical imperative. In our example, it requires of the agent to choose saving Alice, because the goal advances her. There may be several actions advancing the same number of agents, in which case the agent can choose freely (or randomly) amongst them. One could also take Kant to imply a second condition to the meritorious principle, to prevent as many people being negatively affected by circumstances as possible. In the current example, both conditions would lead to the same result.

\subsection{Unhandled case: Not using your talent}

As a final example, consider the following situation: Bob has the talent to become a great artist. However, he wonders whether it is permissible to just be lazy and enjoy life instead of working hard to improve himself.
Strictly speaking Bob is not working to anyone's disadvantage by being lazy and thus the definitions of `being treated as a means' advanced above will not cover this example. As the goal of enjoying life and the goal of making art both benefit Bob, the meritorious principle also cannot be used to make the distinction. What Kant says is that laziness could be consistent with the preservation of humanity but does not harmonize with its advancement. He also writes that a rational being necessarily wills that all his capacities are developed. However, it is not clear to us what constitutes the advancement of humanity beyond the sheer feeling of happiness. The example is further complicated by the fact that Kant says that this is a duty one has towards oneself, not others. Therefore, it would be inappropriate to solve this case by introducing others into the model that would benefit from Bob becoming an artist.

In the current formalization, we have no means to represent the relevant aspects that render laziness impermissible and becoming an artist permissible for the right reasons. We thus take this example to showcase a limitation of our treatment of Kant's ethics, and leave a formalization that could capture this last example for further research.

\section{Implementation within the HERA framework}

The formalization of Kant's second formulation of the categorical imperative has been implemented within the Hybrid Ethical Reasoning Agent software library (short: HERA).\footnote{The HERA software is available from \url{http://www.hera-project.com}. It is fully implemented in Python and can be installed via the PyPI repository (package name: {\tt ethics}).} The general goal of the HERA project is to provide theoretically well-founded and practically usable logic-based machine ethics tools for implementation in artificial agents, such as companion robots with moral competence \cite{LB17}. The core of HERA consists of a model checker for (Kantian) causal agency models. Thus, the situations the agent can reason about are represented in terms of models, and ethical principles like the categorical imperative are implemented as (sets of) logical formulae.
To showcase the use the categorical imperative from a Python program, Listing \ref{listing:case1} reconsiders a representation of the suicide case.

\begin{lstlisting}[language=json, caption=A sample JSON encoding of the
suicide case., label=listing:case1, captionpos=b]{}
{
"actions": ["suicide"],
"background": [],
"consequences": ["dead"],
"patients": ["Bob"],
"mechanisms": {"dead": "suicide"},
"affects": {"suicide": [["Bob", "+"]],
"dead": []},
"goals": {"suicide": ["dead"]}
}
\end{lstlisting}

The workflow for using HERA requires to first generate a causal agency model like the one in Listing \ref{listing:case1}. Given such a model, arbitrary logical formulae can be checked for being satisfied or not by this model. This way, the conditions of ethical principles like the Kantian categorical imperative as defined in Definition \ref{def:categorical_imperative} can be checked for satisfaction.

To support the usage of the HERA library, the logical formulae to be checked for ethical principles already included in HERA are encapsulated into prepared classes. Listing \ref{listing:interaction} shows a sample interaction.
The first three commands load the implementations of two syntactical entities of the logical language (the predicates {\tt Means} and {\tt End}), the causal agency model from the {\tt semantics} package, and the categorical imperative using Reading 1 of `being treated as a means' from the {\tt principles} package. The third command loads the suicide example and sets the external variable $suicide$ to the value \emph{True}. This way, the $suicide$ action is chosen in the situation, and the truth values of the consequence variables can be evaluated the way explained in Section \ref{sect:causal_agency_models}. In the concrete case, \emph{True} will be assigned to the variable $dead$. The fourth command asks whether, in the resulting situation, Bob is used as a means according to Reading 1 (see Definition \ref{def:treated_as_means1}). The answer is \emph{True}, because Bob is affected by the action ($suicide$) and the action is a but-for cause of Bob's goal ($dead$). The fifth command asks if Bob is used as an end. This query returns \emph{False}, because Bob is not affected by the goal (see Section Example 1: Suicide). All in all, the action is not permissible according to the categorical imperative, and the output of the last command is accordingly.

\begin{lstlisting}[language=python, basicstyle=\small, caption={A sample interaction with the Python package {\tt ethics}, which we develop and maintain as the standard implementation of HERA.}, label=listing:interaction, captionpos=b, escapeinside={(*}{*)}]{}
from ethics.language import Means, End
from ethics.semantics import CausalModel as cm
from ethics.principles import KantianHumanityPrinciple as ci
m = cm("suicide.json", {"suicide": True})
m.models(Means("Reading-1", "Bob"))
(*{\bf output:} {\it True}*)
m.models(End("Bob"))
(*{\bf output:} {\it False}*)
m.evaluate(ci)
(*{\bf output:} {\it False}*)
\end{lstlisting}

\section{Conclusion}
We have shown proof of principle how Kant's second formulation of the categorical imperative can be formalized and implemented computationally. The strict duties towards yourself and others are defined, given goals, structural equations, and the affects relation. To define permissibility according the categorical imperative, we have defined `being treated as an end', and we formalized two readings of `being treated as a means' that meet different intuitions about this concept. The formalization deals well with Kant's own examples of strict duties. We were also able to partly deal with Kant's wide duties by defining an additional condition that requires agents to maximize the number of persons being treated as an end.

We envision that the theory will be used as a tool for the comparison of morally relevant aspects of different views on morally delicate cases, thus helping people to have moral discussions. Moreover, we aim at allowing automatic moral judgments in line with Kant in robots such as self-driving cars, care robots, robot companions, and robotic tutors. Our current research investigates whether and under which circumstances Kantian reasoning the way it is presented here is perceived as appropriate for social robots as compared to other types of moral reasoning already defined within HERA.

\end{document}